\documentclass[a4paper,conference]{IEEEtran}

\ifCLASSINFOpdf
\else
\fi
\usepackage{amsmath,amssymb,amsfonts}
\usepackage{algorithmic}
\usepackage{graphicx}
\usepackage{textcomp}
\usepackage{xcolor}
\usepackage{colortbl}
\usepackage{color}
\usepackage{pgfplots}
\usepackage{soul}
\usepackage{subfig}
\usepackage{ctable}
\usepackage{multirow}

\renewcommand{\tablename}{Table }
\renewcommand{\figurename}{Figure }

\IEEEoverridecommandlockouts 
\begin{document}
%
\title{A Digital Image Processing Approach for Hepatic Diseases Staging based on the Glisson's Capsule}


\author{\IEEEauthorblockN{Marco Trombini$^{1}$~\IEEEmembership{Student Member,~IEEE,}, Paolo Borro$^{2}$, Sebastiano Ziola$^{2}$, Silvana Dellepiane$^{1}$~\IEEEmembership{Member,~IEEE,}}

\IEEEauthorblockA{$^{1}$Department of Electrical, Electronic, Telecommunications Engineering and Naval Architecture (DITEN),\\ Università degli Studi di Genova, I-16145, Genoa, Italy\\ Email: marco.trombini@edu.unige.it, silvana.dellepiane@unige.it}
\IEEEauthorblockA{$^{2}$Gastroenterological clinic, IRCCS San Martino Hospital, I-16132, Genoa, Italy.\\
Email: paolo.borro@hsanmartino.it, sebastiano.ziola@hsanmartino.it}
}

%


\maketitle

\begin{abstract}
Due to the need for quick and effective treatments for liver diseases, which are among the most common
health problems in the world, staging fibrosis through non-invasive and economic methods has become of
great importance. Taking inspiration from diagnostic laparoscopy, used in the past for hepatic diseases, in this paper ultrasound images of the liver are studied, focusing on a specific region of the organ where the Glisson's capsule is visible. In ultrasound images, the Glisson's capsule appears in the shape of a line which can be extracted via classical methods in literature. By making use of a combination of standard image processing techniques and Convolutional Neural Network approaches, the scope of this work is to give evidence to the idea that a great informative potential relies on smoothness of the Glisson's capsule surface. To this purpose, several classifiers are taken into consideration, which deal with different type of data, namely ultrasound images, binary images depicting the Glisson's line, and features vector extracted from the original image. This is a preliminary study that has been retrospectively conducted, based on the results of the elastosonography examination.
\newline

\end{abstract}

\section{Introduction}
\label{sec:intro}

Liver diseases are currently a major health problem worldwide due to the increasing incidence and resulting mortality. Hepatic cirrhosis and liver cancer are nowadays among the most common and leading cause of death in the world \cite{asrani2018burden}. More specifically, in the U.S. the mortality due to cirrhosis has been increasing since 2009 \cite{tapper2018mortality}. In this context, the staging of hepatic diseases is therefore of great importance to meet the need for rapid and effective treatments.

The gold standard for liver diseases staging is hepatic biopsy, which however presents several contraindications. Indeed, besides being invasive, this technique may lead complications to patients and only focuses on a small portion of the liver (about 1/50000 of the entire organ \cite{martinez2011noninvasive}). Studies related to serum markers representing the process of hepatic fibrosis have been conducted in order to provide an alternative to biopsy. However, limits of this approach arose, such as the narrow spectrum of disease severity in the included patients. Furthermore, the results correspond to an indirect assessment rather than to direct quantification of the staging of the disease \cite{talwalkar2008magnetic,afdhal2004evaluation}. As a widely used alternative to biopsy, elastosonography provides a quantitative survey on the disease stage by measuring the organ stiffness. Fibroscan Transient Elastography (TE) \cite{de2008transient} and Shear Wave Elastography (SWE) \cite{sarvazyan1998shear} are the most used technologies for this aim but may depend on the operator's experience and therefore they may lead to inconsistent results. Furthermore, such techniques can be affected by pathological conditions such as obesity and liver congestion \cite{perazzo2015factors}.
Another approach to the staging of hepatic diseases relies on diagnostic imaging. Several studies have been conducted using both Magnetic Resonance (MR) and Computerized Tomography (CT) images \cite{talwalkar2008magnetic,pickhardt2016accuracy,kayaalti2012staging}. Even though they provide satisfactory results, such imaging techniques are too expensive or invasive to be performed several times and cannot represent an accessible alternative to hepatic biopsy.

On the contrary, ultrasound imaging is non-invasive and inexpensive and therefore can easily and without contraindications be used. In \cite{meng2017liver}, an approach based on transfer learning is proposed. The five stages of liver fibrosis are re-organized in three (absent, mild, and severe level) by referring to the Ishak score \cite{shiha2011ishak}. The whole image of the organ is processed and relevant features for the classification are extracted via a pre-trained model, requiring a huge quantity of data.

In the present work, taking inspiration from an old diagnostic technique, the aim is to prove that relevant features for the staging of hepatic diseases lie in a specific region of the organ. In particular, in the past, liver diseases were identified via a laparoscopy in the abdominal region, in order to visualize the surface of the fibrous capsule covering the entire organ, known as Glisson's capsule. On one hand, an irregular surface is symptomatic of a diseased organ while, on the other hand, a smooth surface certifies a low level of fibrosis.

In ultrasound images, the Glisson's capsule appears in the shape of a line which will be hereinafter referred to as the Glisson's line. A bright and continuous line represents a smooth surface while a dashed and dark line means that the Glisson's capsule is irregular on the surface.

In \cite{gemme2018quantification} a Region Of Interest (ROI) of the liver is taken into consideration, the Glisson's line is automatically detected and features are consequently extracted and classified in two classes (referring to the Metavir score \cite{shiha2011ishak}, no disease F0-F1 and disease F2-F3-F4) using a Support Vector Machine (SVM) classifier. However, the available dataset is unbalanced (more healthy subjects than impaired patients) and therefore an erroneous tracking of the line in impaired patients did not significantly affect the accuracy of the classification. In this work a homogeneous dataset is used. The innovative aspect of this study is to use a combination of standard image processing techniques and Convolutional Neural Network (CNN) approaches to give evidence supporting the idea that the Glisson's capsule surface encompasses a great informative content related to hepatic diseases. Features are indeed extracted as in \cite{gemme2018quantification} but classified using a Multi-layer Neural Network (MLNN). The results are then compared with the ones obtained from the processing and classification via a CNN. Two CNNs are used in order to process only the ultrasound images and then the ultrasound images along with the extracted lines. If the Glisson's line is correctly detected, the accuracy of the classification improves when both ultrasound and line images are considered. The joint used of classic image processing (for the line extraction) and CNNs provides a first evidence that a huge informative content related to hepatic diseases lies in the Glisson's line. This conjecture is then verified by the fact that the most accurate classification is given by the concatenation of MLNN on the extracted features and the CNN.

The paper is structured as follows: in Section \ref{sec:1} the dataset is presented, along with the the method for features extraction and the Neural Networks (NNs) architectures. Section \ref{sec:2} contains the results of the classification into two (absent disease and disease), three (absent, mild, and severe disease) and five (Metavir score stages) classes. Such results are discussed and compared with similar works and. To conclude, a glimpse into a future development of this study is provided.
\section{Materials and Methods} \label{sec:1}

\begin{figure}[htb]

\begin{minipage}[b]{1.0\linewidth}
  \centering
  \centerline{\includegraphics[width=9cm]{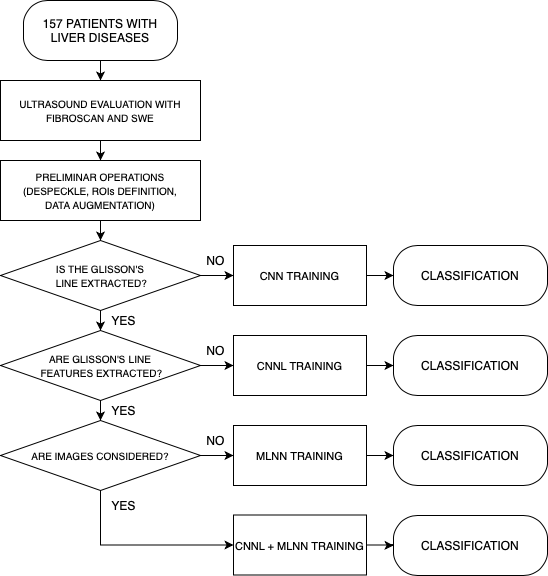}}
\end{minipage}

\caption{Method flow chart. Patients consecutively recruited underwent classical ultrasound evaluation with Fibroscan and SWE. Then the processing phase on the ultrasound images begins. On the basis of the type of input data an extracted features, a different classifier is adopted.}
\label{fig:Method Flow}
\end{figure}

This section presents the available data and the techniques used for the analysis. The flowchart of the present study is reported in Figure \ref{fig:Method Flow}. First of all, ground truth data are obtained, following the medical examination. Then, data are pre-processed and prepared for training the different classifiers. The model is chosen depending on the use of binary images depicting the extracted Glisson's line. In particular, a CNN is used to process only the ultrasound images. Then, ultrasound images along with binary images of the Glisson's line are processed with another CNN, which is indicated as CNNL. Peculiar features along the Glisson's line are obtained similarly to the work in \cite{gemme2018quantification}, and classified via a MLNN. As last alternative, the CNNL and MLNN models are concatenated. 

\subsection{Available data} \label{ssec:data}

This study has been retrospectively conducted  on a population of 157 subjects, with 92 male and 65 female patients. The average age of the whole population is $57.84 \pm 13.39$ years and age ranging from 20 to 92.

Patients presented several types of hepatic disease, the principal are non-alcoholic fatty liver disease (NAFLD), hepatitis C virus (HCV), hepatitis B virus (HBV), autoimmune hepatitis (AIH), and primary biliary cholangitis (PBC).

The image dataset consists of 157 ultrasound liver images acquired by scanning the epigastrium with linear probe, focusing on the anterior border of the third segment of the liver in order to highlight the Glisson's capsule. The multipurpose LOGIQ S8 XDclear 2.0 system was used for ultrasound scanning, performed by a single operator with extensive experience in liver evaluation.


Ground truth data of the disease level consists of the classification provided by the hepatic elastographic exam, according to the Metavir score. Hepatic disease stages are obtained from both Shear Wave elastography (using the convex broadband probe of the GE LOGIQ S8 ultrasound scanner) and FibroScan elastography, performed by the same expert operator. The final values of both the elastographic exams are established as median of at least ten measurements and expressed in kilopascal (kPa). Only the subjects whose outputs from Shear Wave and FibroScan elastographies are equal have been included in the study. Inter quartile range (IQR) and median ratio is used as reliability criteria: only cases with $ {IQR}/{median} \leq 30 \%$ have been included in this study.

The population is then divided into five classes as follows, on the basis of the stage of the disease: $44$ $F0$ subjects (no disease), $31$ $F1$ patients (mild fibrosis), $35$ $F2$ patients (moderate fibrosis), $20$ $F3$ patients (severe fibrosis) and $27$ $F4$ patients (cirrhosis).

\subsection{Preliminar operations} \label{ssec:PreOp}

In this work, two different approaches are attempted, the former taking into consideration the full ultrasound image, the latter focusing on a small ROI whose central area is covered by the Glisson's line. Full images deliver information that relates not only the area very closed to the Glisson's line, but features are also depending on the regions inside and outside the liver. On the other hand, the Glisson's line is very evident in ROI images, thus yielding to a better detection of its profile, since fewer confounding structures are present. Consequently, features computed on the basis of the extracted line are more accurate than in the other situation.

In both the cases, the data are pre-processed to reduce the speckle noise that typically affects ultrasound images. Then, the size of the dataset in augmented, so that the use of machine learning techniques for the classification is enabled.

\subsection*{Pre-processing}

A filtering operation is performed with the aim of reducing the speckle noise that typically affects ultrasound images. In particular, the SRAD filter (Speckle Reducing Anisotropic Diffusion) \cite{sarode2011reduction} was used. Then, ROI images are obtained by  resizing them to $310 \times 90$ dimension, focusing on the area that contains the Glisson's line.

\subsection*{Data augmentation}

\begin{figure}[htb]

\begin{minipage}[b]{1.0\linewidth}
  \centering
  \centerline{\includegraphics[width=9cm]{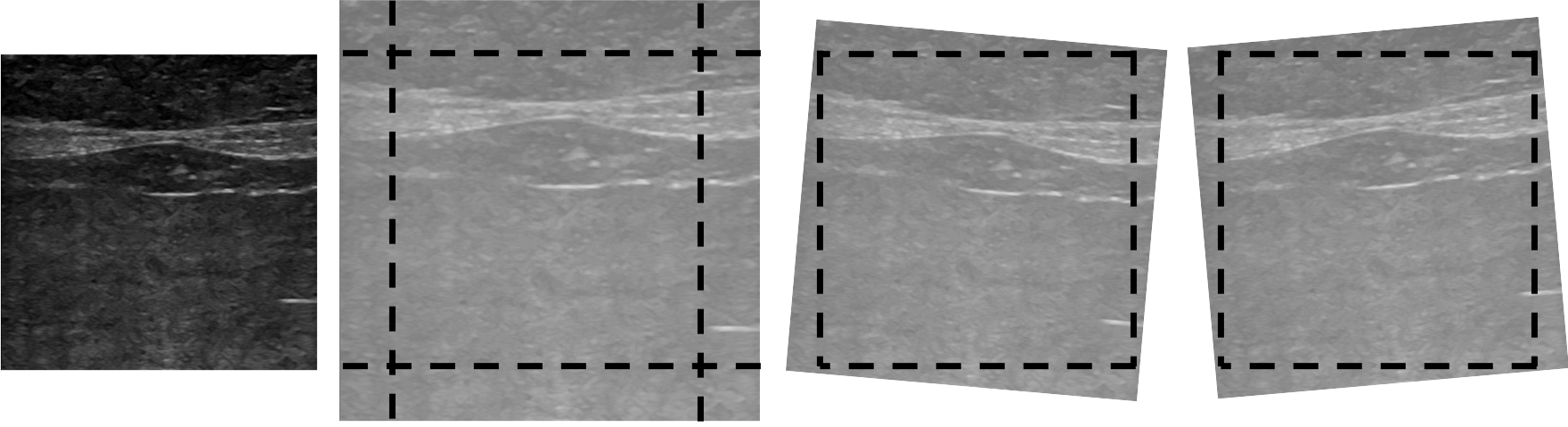}}
\end{minipage}

\caption{Example of data augmentation for images: crop-and-zoom and  small rotation ($\pm$5 degrees) along the probe axis. These small operations allow for obtaining new independent and identically distributed data, without altering too much the informative content, thus providing plausible ultrasound images.}
\label{fig:DataAug}
\end{figure}

The limited size of image dataset is a common problem when applying machine learning methods to the medical field. The same issue was addressed in the work in \cite{meng2017liver} by resorting to data augmentation techniques, thus a similar solution is here used. In particular, in this work, a crop-and-zoom operation or a small rotation along the probe axis are performed on the images, as specified in Figure \ref{fig:DataAug}.

Later, the Glisson's line is consequently detected from the modified images.

\subsection{Line extraction} \label{ssec:LineExtr}

The Glisson's line profile is automatically extracted using the algorithm in \cite{gemme2018quantification}. In particular, after the pre-processing phase devoted to reduce the speckle noise, the contrast in ultrasound images is increased, in order to highlight the Glisson's line profile, which appears bright over the dark background. Then, an edge detection operator (Prewitt operator \cite{shapiro130307963computer}) is used for the identification of the line. Based on such detected line, the features described in Paragraph \ref{ssec:Features} are extracted.

\begin{figure}[htb]

\subfloat[Example of full image and line \label{subfig-2:Full}]{
\centering
\centerline{\includegraphics[width=7cm]{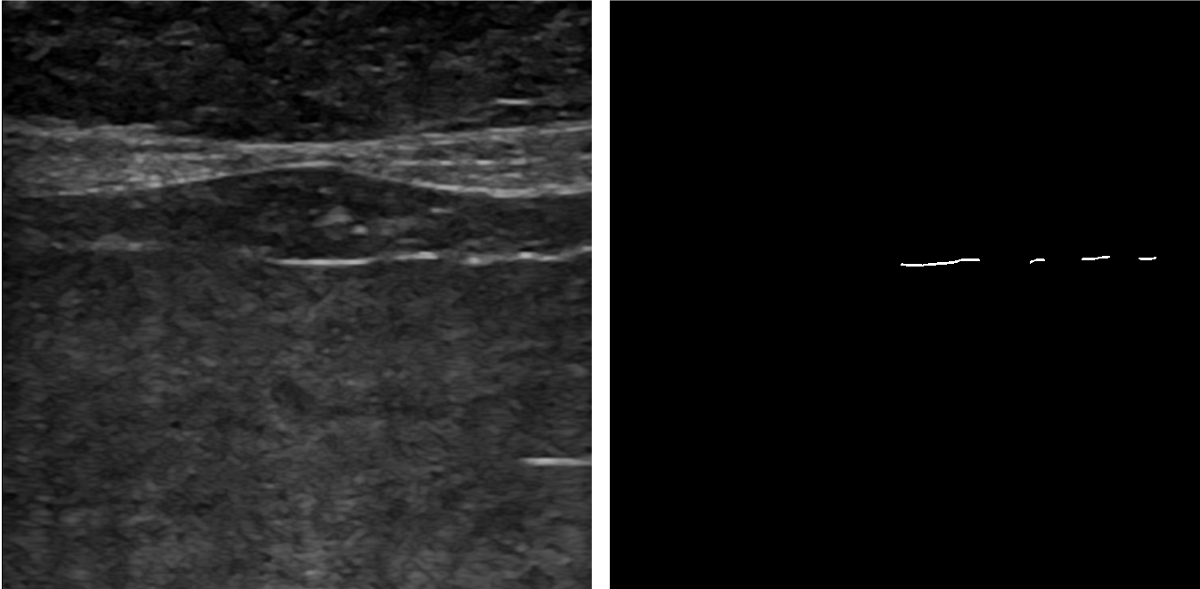}}
}
\\
\subfloat[Example of ROI image and line \label{subfig-2:ROI}]{
  \centering
  \centerline{\includegraphics[width=7cm]{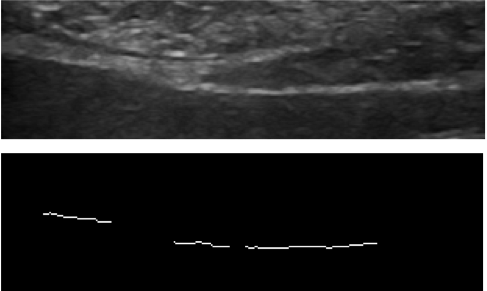}}
}

\caption{Example of Glisson's line extraction from patients' images. Subfigure \ref{subfig-2:Full} depicts the full ultrasound image of a F3 patient along with the corresponding binary image of the Glisson's line. Subfigure \ref{subfig-2:ROI} regards a ROI image of a patient with fibrosis at level F1.}
\label{fig:LineExtraction}
\end{figure}

Figure \ref{fig:LineExtraction} depicts the result of the process, both in the case of full and ROI images. As aforementioned in the previous section, ROI images yield to a better detected line than when considering full images, but other features deriving from the background may be lost.

\subsection{Features} \label{ssec:Features}

Image features are extracted in a neighbourhood of the detected line. In particular, first of all the gray level gradient along the extracted line is calculated, to describe the variation of the grey levels surrounding the line. Mean of gradient and variance of gradient indeed summarise the information regarding the values of the grey levels and their variability. High gradient values and low variability are expected to characterise low stages of the disease. Furthermore, the Glisson's line itself encloses some relevant features, i.e. line continuity and length. They are indicative of the degree of line fragmentation, which is higher in advanced stages of the disease. To conclude, image contrast represents the measure of the local intensity variation.

\subsection{Classifiers} \label{ssec:Classifiers}

The four models for staging hepatic disease based on the available data are here presented.

\subsection*{CNN processing}
\label{ssec:CNN processing}

\begin{figure}[htb]

\begin{minipage}[b]{1.0\linewidth}
  \centering
  \centerline{\includegraphics[width=9cm]{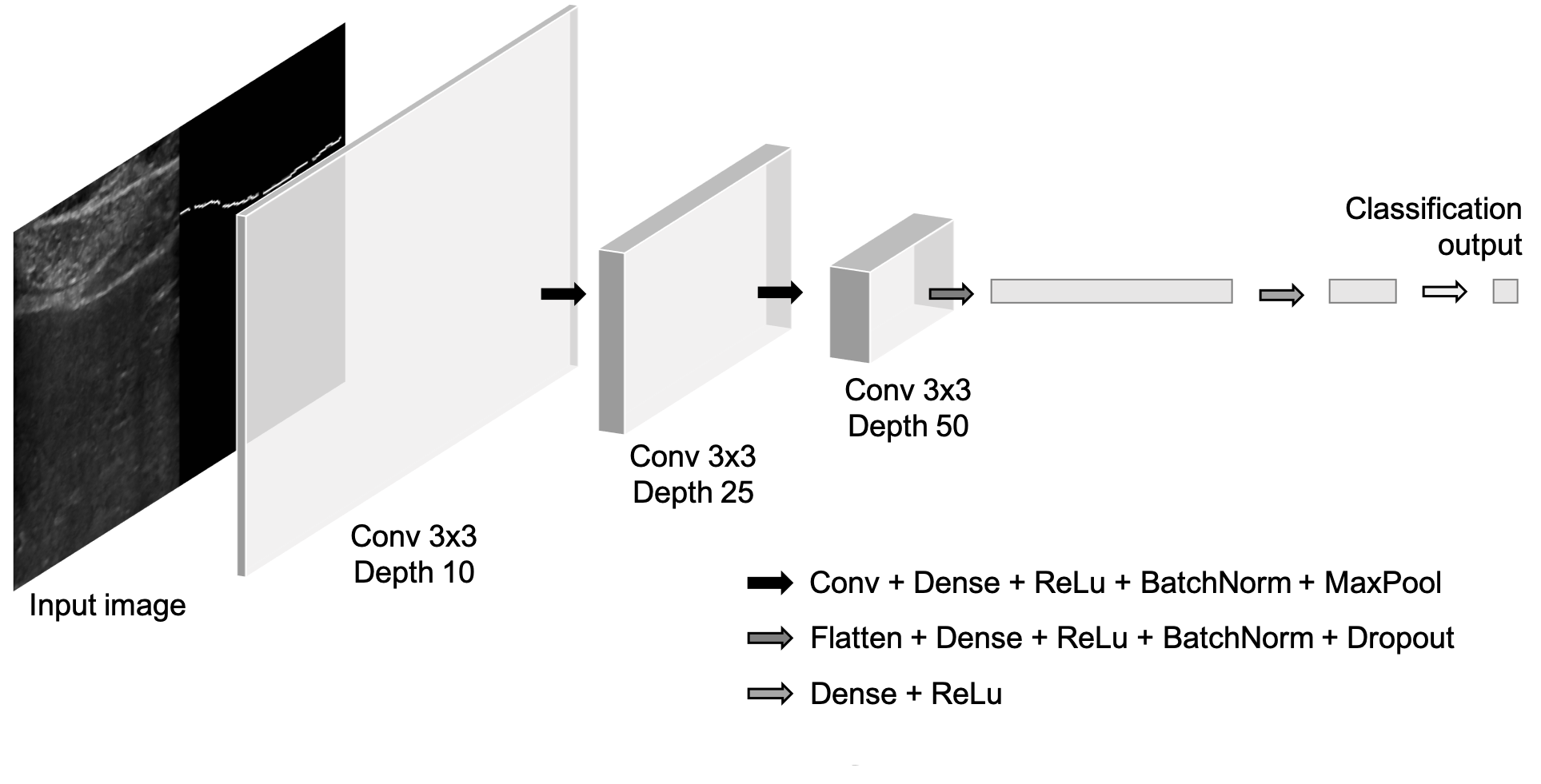}}
\end{minipage}

\caption{Convolutional neural network used for image processing. In the present figure the application to the joint image composed by both ultrasound and binary images is shown. It inherits the same architecture the CNN processing only ultrasound images, but with input data of different size.}
\label{fig:CNN}
\end{figure}

The first approach to the classification was attempted using a CNN. Experimental trials showed that a very deep network is not necessary to process the available data. The reason is that global features are not particularly informative in such ultrasound images, thus not requiring the deep feature extraction granted by deep CNNs. Conversely, low-level and local features have been shown to be particularly informative \cite{gemme2018quantification}, yielding the choice to a simpler and shallower network architecture. The network is described in Figure \ref{fig:CNN}. This CNN was used with different types of input: full ultrasound images alone and together with the Glisson's line image (as in Figure \ref{fig:CNN}), ROI ultrasound images alone and together with the Glisson's line image.

\subsection*{MLNN classifier}
\label{ssec:MLNN}

The conventional approach to extract peculiar features along the Glisson's line proposed in Paragraph \ref{ssec:Features}, yields to an alternative model to the CNN. Similarly to the work \cite{gemme2018quantification}, an algorithm for staging the disease based on the extracted feature was used. However, instead of an SVM classifier, an MLNN was chosen, to enable not only binary classification, but also consider three and five classes. Features referring to ROI images result more accurate than in the full images case, due to the better line detection allowed by focusing on a small area.

\subsection*{Concatenated model}
\label{ssec:Concatenated model}

\begin{figure}[htb]

\begin{minipage}[b]{1.0\linewidth}
  \centering
  \centerline{\includegraphics[width=9cm]{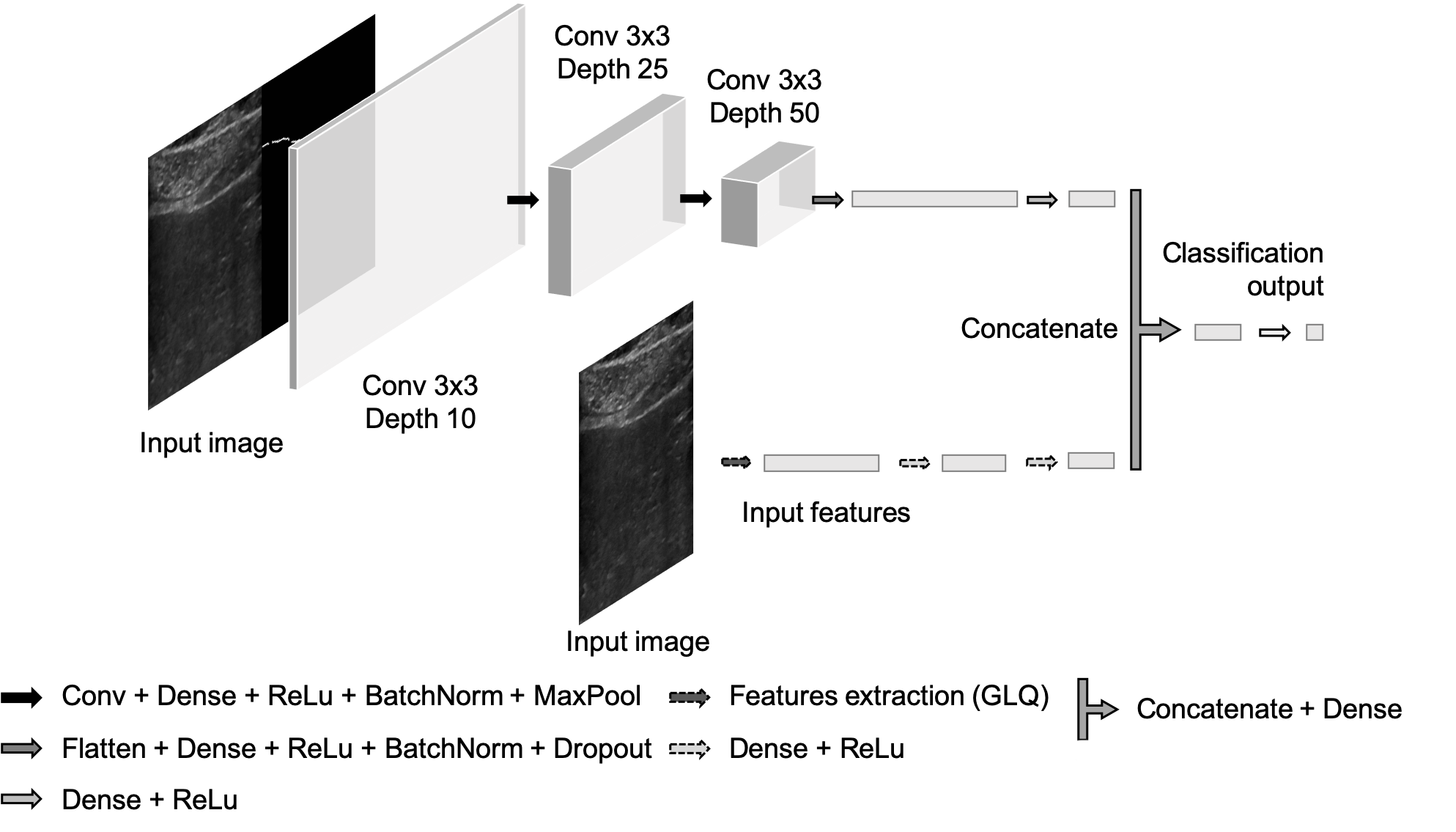}}
\end{minipage}

\caption{Full model obtained by concatenating the CNN and the MLNN. In the upper part, the CNN proposed for analysing joint images containing both ultrasound and binary images is shown. In the lower part, the features extraction process is implicitly included as first step, which provides the features vector to be fed in the MLNN.}
\label{fig:FullModel}
\end{figure}

In order to identify the approach leading to the best result and whether the union of the previously described methods provides a better model, the two processing steps have been concatenated as in Figure \ref{fig:FullModel}. Inputs of the two parts are still disjoint.

\section{Experimental results and discussion}
\label{sec:2}

The available dataset has been classified according to two, three, and five classes. More specifically, when considering two classes, the dataset was divided into \textit{healthy subjects/low disease patients} (stages $F0$ and $F1$) and \textit{impaired patients} (stages $F2$, $F3$, and $F4$); when referring to three classes, the considered groups were \textit{healthy subjects} (stage $F0$), \textit{low/mild level disease patients} (stages $F1$ and $F2$), and \textit{severe stage/cirrhotic patients} (stages $F3$ and $F4$).

The study consisted of four steps:
\begin{itemize}
    \item [1)] CNN to process only the ultrasound images (both ROIs and global);
    \item [2)] CNN to process the ultrasound images (both ROIs and global) together with the lines extracted by the corresponding approach (CNNL);
    \item [3)] features extraction from the ultrasound images (both ROIs and global) and classification via MLNN;
    \item [4)] processing of the ultrasound images (both ROIs and global) together with the extracted lines, using the full model (CNNL + MLNN).
\end{itemize}

All models $1)$, $2)$, $3)$ and $4)$ were independently trained from scratch.

\subsection{Classification results}
\label{ssec:results}

The previously described methods were applied to a set of $628$ elements (where with \textit{element} it is meant a combination of ultrasound images, binary images, and features vector, depending on the used model).

The dataset was split into training, validation, and test sets with percentage of $80\%$, $10\%$, and $10\%$ respectively, allowing for 10-fold cross-validation and in order to control for over-fitting.

The performance of the suggested approaches was evaluated using the following metrics:
\begin{itemize}
    \item Overall accuracy: \[Acc = \frac{|\text{correct \  predictions}|}{|\text{predictions}|};\]
    \item Mean Absoulte Error of the predicted classes: \[MAE = \displaystyle \frac{1}{N_{t}}\sum_{i=1}^{N_{t}} |cc_i - pc_i|,\] where $cc_i$ is the correct class number of $i^{-th}$ element, $pc_i$ is the predicted class number of $i^{-th}$ element, and $N_{t}$ is the size of the test set..
\end{itemize}{}

Randomness was accounted by performing cross-validation on 25 permutations and the average overall accuracy and MAE are summarised in Tables \ref{tab:results table 1} and \ref{tab:results table 2} with respect to all the trials and referring to full and ROI images respectively.

\begin{table}[h!]
\caption{Overall Accuracy and Mean Absolute Error for each model, full image data.}
\centering
\begin{tabular}{|c|c|c|}
\hline
 & \multicolumn{2}{c|}{\textbf{Two classes}}\\ \hline
 & CNN & CNNL \\ \hline
Acc & 73,13$\%$ $\pm$\ 0,02 & 75,94$\%$ $\pm$ \ 0,04 \\ \hline
MAE & 0,27 $\pm$ \ 0,02 & 0,24 $\pm$ \ 0,04 \\ \hline
 & MLNN & CNNL+MLNN \\ \hline
Acc & 94,69$\%$ $\pm$ \ 0,01 & 95,94$\%$ $\pm$ \ 0,02 \\ \hline
MAE & 0,05 $\pm$ \ 0,01 & 0,04 $\pm$ \ 0,02 \\ \hline
 & \multicolumn{2}{c|}{\textbf{Three classes}}\\ \hline
 & CNN & CNNL \\ \hline
Acc & 60,94$\%$ $\pm$ \ 0,12 & 67,19$\%$ $\pm$ \ 0,03 \\ \hline
MAE & 0,39 $\pm$ \ 0,01 & 0,32 $\pm$ \ 0,04 \\ \hline
 & MLNN & CNNL+MLNN \\ \hline
Acc & 81,25$\%$ $\pm$ \ 0,02 & 85,93$\%$ $\pm$ \ 0,02 \\ \hline
MAE & 0,19 $\pm$ \ 0,01 & 0,14 $\pm$ \ 0,02 \\ \hline
 & \multicolumn{2}{c|}{\textbf{Five classes}}\\ \hline
 & CNN & CNNL \\ \hline
Acc & 56,25$\%$ $\pm$ \ 0,11 & 62,50$\%$ $\pm$ \ 0,03 \\ \hline
MAE & 0,44 $\pm$ \ 0,02 & 0,38 $\pm$ \ 0,04 \\ \hline
 & MLNN & CNNL+MLNN \\ \hline
Acc & 79,68$\%$ $\pm$ \ 0,01 & 84,38$\%$ $\pm$ \ 0,13 \\ \hline
MAE & 0,20 $\pm$ \ 0,04 & 0,16 $\pm$ \ 0,03 \\ \hline
\end{tabular}
\label{tab:results table 1}
\end{table}

\begin{table}[h!]
\caption{Overall Accuracy and Mean Absolute Error for each model, ROI image data.}
\centering
\begin{tabular}{|c|c|c|}
\hline
 & \multicolumn{2}{c|}{\textbf{Two classes}}\\ \hline
 & CNN & CNNL \\ \hline
Acc & 74,69$\%$ $\pm$ \ 0,02 & 77,50$\%$ $\pm$ \ 0,04 \\ \hline
MAE & 0,23 $\pm$ \ 0,02 & 0,23 $\pm$ \ 0,04 \\ \hline
 & MLNN & CNNL+MLNN \\ \hline
Acc & 96,25$\%$ $\pm$ \ 0,01 & 97,50$\%$ $\pm$ \ 0,01 \\ \hline
MAE & 0,04 $\pm$ \ 0,02 & 0,03 $\pm$ \ 0,02 \\ \hline
 & \multicolumn{2}{c|}{\textbf{Three classes}}\\ \hline
 & CNN & CNNL \\ \hline
Acc & 64,38$\%$ $\pm$ \ 0,03 & 72,19$\%$ $\pm$ \ 0,01 \\ \hline
MAE & 0,36 $\pm$ \ 0,02 & 0,27 $\pm$ \ 0,01 \\ \hline
 & MLNN & CNNL+MLNN \\ \hline
Acc & 87,81$\%$ $\pm$ \ 0,02 & 90,31$\%$ $\pm$ \ 0,04 \\ \hline
MAE & 0,12 $\pm$ \ 0,01 & 0,10 $\pm$ \ 0,03 \\ \hline
 & \multicolumn{2}{c|}{\textbf{Five classes}}\\ \hline
 & CNN & CNNL \\ \hline
Acc & 52,50$\%$ $\pm$ \ 0,02 & 63,13$\%$ $\pm$ \ 0,01 \\ \hline
MAE & 0,48 $\pm$ \ 0,01 & 0,34 $\pm$ \ 0,01 \\ \hline
 & MLNN & CNNL+MLNN \\ \hline
Acc & 85,94$\%$ $\pm$ \ 0,03 & 88,13$\%$ $\pm$ \ 0,02 \\ \hline
MAE & 0,14 $\pm$ \ 0,02 & 0,12 $\pm$ \ 0,01 \\ \hline
\end{tabular}
\label{tab:results table 2}
\end{table}

\subsection{Discussion}

In both ROI and full images cases, the classification accuracy increases as the number of information processed by the model grows. Similarly, the MAE which each model commits decreases when more detailed data are fed in as input.

However, both accuracy increase and MAE decrease are mostly significant when features extracted on the basis of the Glisson's line are considered. Indeed, even though in MLNN model ultrasound images are not involved, its performance is better than in the case of CNN and CNNL. Then, a small improvement for the classification happens when considering the full model, thus confirming the assumption according to which the Glisson's capsule is responsible for the highest information content related to the hepatic diseases.

Furthermore, when comparing results with respect to full and ROI images, it can be noticed that the classification is more accurate in the latter case. Indeed, when focusing on a small area, the line detection process is more precise than in the full image case, and so are the extracted features. The informative content relying in the background turn to be not only not very significant, but also confounding. One can deduce that contrast is not a very discriminant feature, while the most characterising ones are gradient values and length and continuity of the detected line. This is coherent with the features analysis performed in \cite{gemme2018quantification}.

As for the comparison with other similar works, the two-classes classification via MLNN and MLNN+CNNL provides better results than the ones in \cite{gemme2018quantification}, in both the cases of full images and ROIs. In addition, the outcome in the case of three classes is comparable to the ones in \cite{meng2017liver} even though the approach is different in terms of staging method. No literature dealing with five classes was found.

To conclude, it results that the better the identification of the line, the more accurate the classification. Therefore, a future development of this work will be improving the line detection process. So far, the support of classic image processing techniques to accurately extract the line appeared particularly effective. Alternatively, if a large dataset is available, an automatic detector trained on ground truth data, which consists on blinded manual extractions can be design. Such a detector can be included in an end-to-end version of the presented classifiers, thus yielding to a homogeneous and compact method.

The present study is a preliminary work, due to some limitations i.e. the use of elastometric results as ground truth data rather than the histological examination, and the small sample size. As for the former issue, ground truth staging was indeed provided by Fibroscan and SWE examination when both the methodology were concordant (to address the problem of user-dependance of SWE diagnosis). However, biopsy is the gold standard for staging liver diseases even though its applicability is limited. Hence, a long time is required to collect enough data to test a classification based on the bioptic exam, which is necessary to validate the proposed approach. In the future, also the sample size problem will be addressed.

Some staging techniques, included Fibroscan and SWE, may depend also on pathological conditions, such as obesity and liver congestion, thus affecting the goodness of the diagnosis. On the contrary, since the proposed study is focused on morphological criteria, based on the Glisson's line, any modifications on the liver status, deriving from pathological conditions, does not influence the results of the current approach. Hence, a further investigation on discordant results from Fibroscan an SWE will be conducted as future development.

\section{Conclusion}
\label{sec:3}

In this paper, a study related to a new diagnostic approach for hepatic diseases, based on ultrasound images, was presented. The proposed method has several positive aspect in terms of applicability since it is low costs, rapid and easy to perform, safe, and uses objective and non-operator dependent criteria.

The outcome of the present study proves that a great informative potential relies in the so called Glisson's line. Ground truth data comes from the validated technique of elastosonography and different approaches have been compared and discussed. 


Further analysis, involving more patients and possibly multicenter, are required to validate the results of this preliminary study, which however are promising as pilot. The reliability of the present work will be studied via \textit{test-retest} and \textit{intra/inter rater} evaluations, in order to make it more statistically relevant.

\end{document}